\documentclass[journal]{IEEEtran}
\usepackage{microtype}                 
\PassOptionsToPackage{warn}{textcomp}  
\usepackage{tikz}
\usepackage{comment}
\usepackage{amsmath,amssymb} 
\usepackage{color}
\usepackage{lineno,hyperref}
\usepackage{comment}
\usepackage{bbm}
\usepackage{amsmath, xparse}
\usepackage{here}
\usepackage{xcolor}
\usepackage{subfig}
\usepackage{bm}
\usepackage{multirow}
\usepackage{ctable} 
\usepackage{arydshln}
\usepackage{algorithm}
\usepackage{amsfonts}
\usepackage{bbding}
\usepackage{pifont}
\usepackage{wasysym}
\usepackage{amssymb}
\usepackage[noend]{algpseudocode} 
\usepackage{algorithmicx}

\newcommand{\qin}[1]{{\textcolor{black}{#1}}}

\ifCLASSINFOpdf
\else
\fi
\begin{document}
%
\title{Teacher-Student Network for 3D Point Cloud Anomaly Detection with Few Normal Samples}
%
%
%

\author{Jianjian Qin, Chunzhi Gu,
        Jun Yu, 
        and~Chao~Zhang
\thanks{Manuscript received XX, xx; revised XX. (Corresponding author: C. Zhang.) }
\thanks{J. Qin, C. Gu, and C. Zhang* are with the School of Engineering, University of Fukui, Fukui, Japan (e-mails: qjd21006@u-fukui.ac.jp; gchunzhi@u-fukui.ac.jp; zhang@u-fukui.ac.jp).}
\thanks{J. Yu is with Institute of Science and Technology, Niigata University, Japan (e-mail: yujun@ie.niigata-u.ac.jp).}
}

\maketitle

\begin{abstract}

Anomaly detection, which is a critical and popular topic in computer vision, aims to detect anomalous samples that are different from the normal (i.e., non-anomalous) ones. The current mainstream methods focus on anomaly detection for images, whereas little attention has been paid to 3D point cloud. In this paper, drawing inspiration from the knowledge transfer ability of teacher-student architecture and the impressive feature extraction capability of recent neural networks, we design a teacher-student structured model for 3D anomaly detection. Specifically, we use feature space alignment, dimension zoom, and max pooling to extract the features of the point cloud and then minimize a multi-scale loss between the feature vectors produced by the teacher and the student networks.
Moreover, our method only requires very few normal samples to train the student network due to the teacher-student distillation mechanism. Once trained, the teacher-student network pair can be leveraged jointly to fulfill 3D point cloud anomaly detection via the calculated anomaly score. To the best of our knowledge, our method is the first attempt to realize anomaly detection for point cloud with few samples. Extensive experimental results and  ablation studies quantitatively and qualitatively confirm that in the case of training with very few samples, our model can achieve higher performance compared with the state of the arts in 3D anomaly detection.

\end{abstract}

%
\IEEEpeerreviewmaketitle

\section{Introduction}
\label{introduction}
Anomaly detection (AD), which refers to distinguishing patterns in data that do not satisfy expected behavior, plays a key role in diverse research areas such as industrial inspection \cite{de2022hybrid, wang2021student_teacher, ko2022new, roth2022towards, li2021cutpaste} and biomedical signal processing \cite{adhikary2022dynamic, khalifa2021review, zhao2021anomaly,pinaya2022unsupervised,chen2022utrad}. Specifically, because of the clear explainability and explicit location of anomalies in real-world applications, current AD techniques have focused on detecting visual anomalies in images (e.g., damages on the surface of an object).

\begin{figure}[tb]
 \centering
 \includegraphics[width=0.5\textwidth]{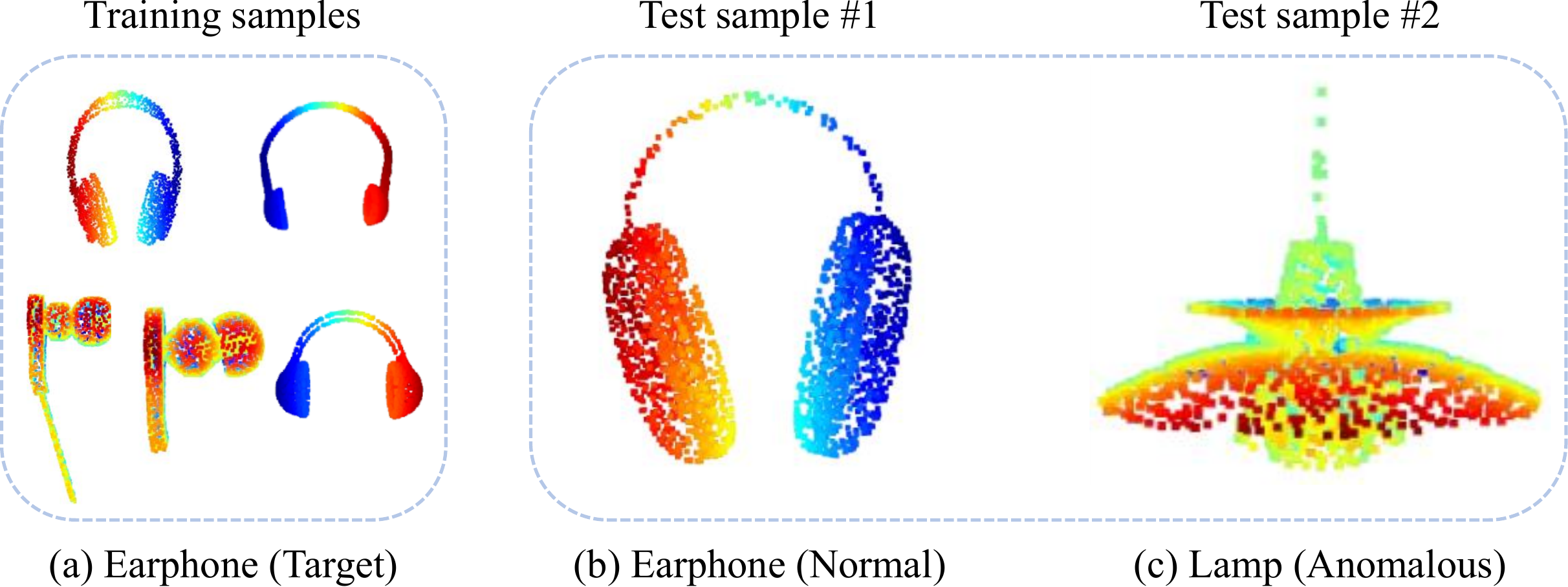}
 \caption{An example of anomaly detection for 3D point cloud. Categories that differ from the training data are defined as anomalies. Our method enables training with few (e.g., five) samples of the target normal category shown in (a) to correctly detect the normal (b) and anomalous (c) samples during testing.}
 \label{fig:1}
\end{figure}

Compared with 2D representations such as images, 3D data generally offers a more comprehensive understanding for the real world \cite{collins2022abo, yi2017large, liao2022kitti, bergmann2021mvtec}. Especially, with the rapid development of 3D sensors (e.g., LiDAR), dealing with the easily accessible 3D data has received growing attention, which necessitates AD techniques to provide quality inspection in various manufacturing scenarios. In the context of AD for 3D point cloud, one of the core problems lies in how to effectively detect the point cloud for undesired (i.e., anomaly) objects given a target normal object class. Fig. \ref{fig:1} shows an example. Given the assumed object category of ``Earphone" (Fig. \ref{fig:1}(a)), the AD algorithm is expected to distinguish the anomalous test samples with different category labels (e.g., ``Lamp" in Fig. \ref{fig:1}(c)) while being able to recognize the normal category (Fig. \ref{fig:1}(b)). Similar out-of-category problem settings can be found in \cite{akcay2018ganomaly,kimura2020adversarial,masuda2021toward}. To achieve this, one straightforward method is to directly apply 2D-based AD approaches to 3D point cloud. However, as pointed out in \cite{bergmann2022anomaly}, the simple adoption of 3D counterparts for the 2D ones hardly results in promising performance because 3D data is naturally unordered and inherently  noisier. 

\begin{figure*}[h]
 \centering
 \includegraphics[width=1\textwidth]{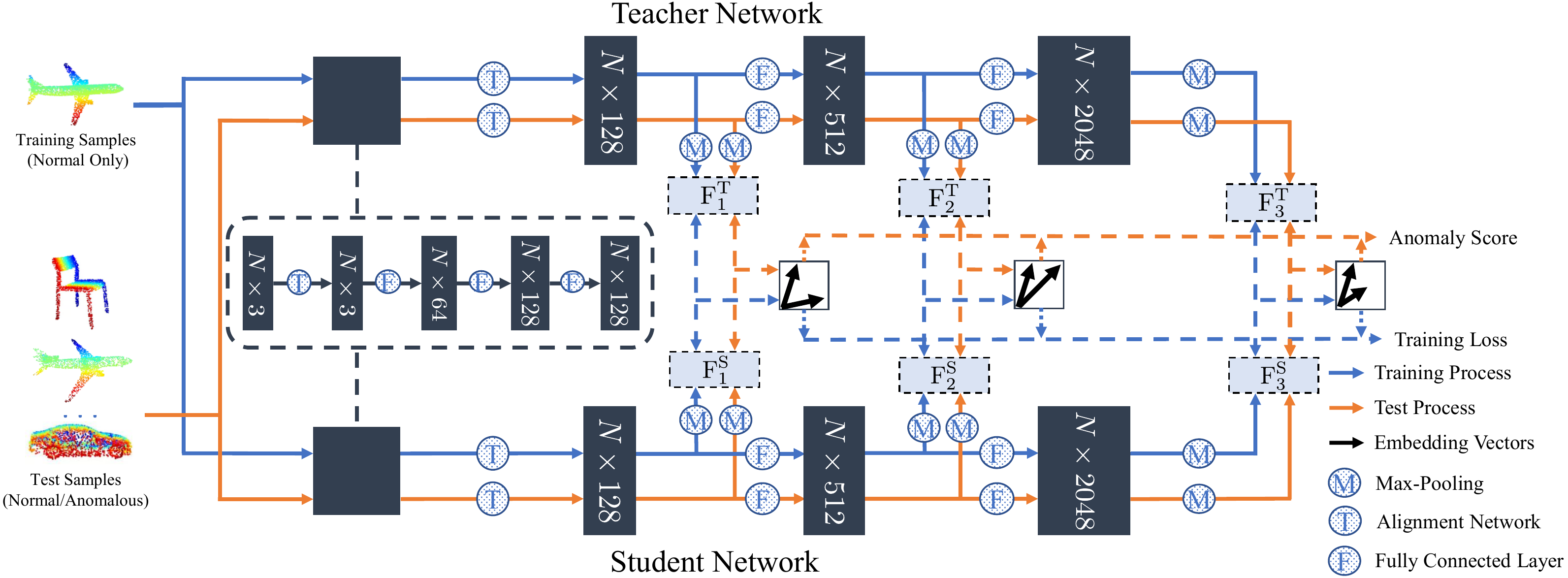}
 \caption{Method overview of our teacher-student AD framework for 3D point cloud. In the training stage, the student network is trained on a specific normal sample category to learn multi-scale features to imitate the output of the corresponding teacher network. During the test stage, the degree of anomaly is calculated according to the similarity between the feature vectors of the teacher and the student.}
 \label{fig:2}
 \end{figure*}

A pioneering work by \cite{masuda2021toward} first attempts to resolve AD for 3D objects with a Variational Auto-Encoder (VAE) framework. They first train the VAE with a large set of normal 3D samples presented by point clouds. Since the anomalous samples are unseen to the trained VAE, the poorly reconstructed test samples indicate abnormality. Despite the conceptual novelty and the satisfactory detection accuracy, this reconstruction-based method requires a large number of normal samples for training to empower the VAE with decent reconstruction capability. However, obtaining sufficient training samples for all classes in the real world involves prohibitively large efforts especially when the number of samples is limited. In addition, the selection of a different normal category will demand a retraining procedure with heavy computational expense, which lowers its usability and practicality.

To address the issues described above, in this paper, we propose a new AD method for 3D point cloud data based on the teacher-student network architecture. Our key idea is to leverage the teacher-student distillation mechanism such that the student network can specifically model the distribution for normal samples. It is worth mentioning that with the pretrained teacher network extracting all of the necessary features for the point cloud, optimizing our student network only requires very few normal samples to implement anomaly detection for specific categories. Such a beneficial property contributes to reducing the computational cost of retraining induced by the change of the normal category compared with the existing AD methods. Experiments conducted on a widely known point cloud dataset demonstrate superior anomaly detection performance with only few normal samples for training.

In summary, our paper includes the following main contributions:

\begin{itemize}
\item We propose a teacher-student framework embedded with the PointNet architecture to address the task of anomaly detection for point cloud, which is among the first to realize 3D anomaly detection for general objects. Furthermore, our framework can be trained with very few normal samples to realize this task.

\item We design a training strategy by introducing a multi-scale loss to seamlessly merge the pooled features at different layers to guide the student network to learn quality features for detection. 

\item We perform extensive experiments and ablation studies against two state-of-the-art models to qualitatively and quantitatively demonstrate the effectiveness of our method in achieving superior detection accuracy with only few training samples.

\end{itemize}


\section{Related Works}
\label{Related Works}

In this section, we first discuss the current mainstream 2D AD approaches, and then introduce some AD methods for 3D point cloud. Then, we review some teacher-student structured models for AD. Finally, we provide a comprehensive introduction to the AD techniques which use a few sample training strategy.\par

\subsection{Anomaly Detection for 2D Images}
The mainstream AD methods \cite{schlegl2017unsupervised,akcay2019ganomaly,sabokrou2018adversarially,sabokrou2017deep,sabokrou2019avid, pourreza2021g2d} have been advanced in the field of 2D images. \cite{schlegl2017unsupervised} leveraged a deep convolutional generative adversarial network to learn a manifold of normal anatomical variability to make use of insufficient anomalous samples. \cite{sabokrou2018adversarially} proposed a general framework with paired reconstructor and discriminator models for novelty detection and one-class classification, respectively. However, images often do not reflect the fine geometric structures of objects well due to depth ambiguity. Since these methods are specifically designed to handle 2D representations, applying them directly on 3D AD tasks can be less practical. Therefore, AD techniques for 3D data need to be studied to explore better solutions for real-world applications.

\subsection{Anomaly Detection for 3D Point Cloud}
It is worth mentioning that AD for 3D point cloud has been explored very little to date. To the best of our knowledge,  \cite{masuda2021toward},  \cite{bergmann2022anomaly}, and \cite{floris2022composite} constitute the only explorations of this field. \cite{bergmann2022anomaly} first adopted the teacher-student network to compute the regression error as the anomaly score for 3D anomaly detection, and the teacher network is pretrained by reconstructing local receptive fields. This work focuses on point-wise anomaly detection, and the feature extracted from the point cloud with color information is dense and represented in the form of 3D descriptors. As for the dataset they used (MVTec 3D-AD dataset \cite{bergmann2021mvtec}), since the object is located within a limited region and the cameras are well calibrated, the effect induced by the geometric transformation of the object can be largely disregarded. On the other hand, at this point, our problem setting is more challenging because the extracted features should be invariant to rotations, translations, and scale change, which is the main reason for using PointNet as the feature extractor in this paper. \cite{floris2022composite} introduced a composite layer  as a convolutional operator to extract and compress the spatial information from point position for AD. \cite{masuda2021toward} proposed a VAE-based AD framework for 3D point cloud. This method handles AD for general objects represented by 3D points. However, reconstruction often incurs huge computational expense. Additionally, achieving a quality reconstructing capability generally involves a large number of training samples, which makes it less applicable in the real world. Contrarily, since we exploit the teacher-student distillation scheme, our method shows less reliance on sample number compared with reconstruction-based methods. 

\subsection{Anomaly Detection with Teacher-Student Networks }
Teacher-student-based methods \cite{bergmann2020uninformed,wang2021student_teacher,salehi2021multiresolution,kumari2022adaptive,deng2022anomaly,shen2021real} have been extensively studied  in the field of AD, particularly focused on 2D images. 
\cite{bergmann2020uninformed} proposed a teacher-student framework for unsupervised anomaly detection with discriminative latent embeddings to efficiently construct a descriptive teacher network. 
In addition, a dedicated network design can exploit more information from the rich features.
\cite{wang2021student_teacher,salehi2021multiresolution} leveraged both coarse and fine levels of features by incorporating a multi-scale training scheme into the teacher-student framework. Despite the remarkable performance, these methods are limited to handling AD for 2D images only. 

Later efforts \cite{rudolph2023asymmetric,yamada2022reconstructed,zhou2022pull,kumari2022adaptive,deng2022anomaly} then extended the teacher-student framework to explore broader applications to other forms of data. \cite{rudolph2023asymmetric} formulated an asymmetric teacher-student network with a bijective normalizing flow as the teacher to ensure that the distance of anomalies in the feature space is sufficiently large. \cite{yamada2022reconstructed} proposed a framework with two pairs of teacher-student networks equipped with a discriminative network to learn accurate anomaly maps. The teacher-student model by \cite{zhou2022pull} addresses the high false detection probability imposed by subtle defects by maximizing pixel-wise discrepancies for anomalous regions. \cite{kumari2022adaptive} proposed an adaptive framework to understand scene dynamics from audio-visual data in a hybrid fusion manner. \cite{deng2022anomaly} proposed a reserved distillation method by using a one-class bottleneck embedding (OCBE) module to boost the discrimination capability. Overall, a crucial problem for these methods is that a large number of normal samples are required for training, yet it is not always feasible to acquire sufficient training data. Moreover, since they involve retraining procedures to tackle another normal category assumption, retraining on all the categories can result in highly expensive computational consumption.

\subsection{Few Samples-based Anomaly Detection}
To alleviate the reliance on large training samples, recent works have shifted to exploring few samples-based AD techniques. Because of the inherent difficulty that a quality generalization ability generally involves more than thousands of samples \cite{lu2020learning,bendre2020learning,wang2020generalizing}, AD with few samples has still been treated very sparsely. Here, we investigated two approaches related to few-samples training in a broader context: the few-shot approach and the data augmentation approach.\par

As for the few-shot based methods, in \cite{sato2022few}, a multilevel VAE is proposed to separate domain-level features from sample-level features for domain generalization with few samples. In \cite{pang2021explainable}, a novel prior-driven model is created to achieve an end-to-end differentiable learning of fine-grained anomaly score by utilizing a small number of labeled anomalies with a Gaussian prior. To address the over-fitting problem when dealing with sparsely labeled data, the authors of \cite{zhou2020siamese} designed a Siamese Convolutional Nerual Network (CNN)-based model with a relative-feature representation scheme. A hierarchical generative model was utilized in \cite{sheynin2021hierarchical} to capture the multi-scale patch distribution of few training images. In \cite{zhang2020few}, a model-agnostic meta-learning model (MAML) is proposed, which can quickly adapt to unseen tasks with an inner loop and an outer loop inside to detect anomalies with few samples.

There are also works to artificially increase the amount of data for solving the few samples problem (i.e., data augmentation). In \cite{zheng2020generic}, they randomly masked out the square regions of an input image, named ``Cutout", to realize automated surface inspection. To overcome the tedious laboring efforts in manual annotation, several generative model-based data augmentation techniques \cite{jain2020synthetic, yun2020automated} have been proposed. \cite{jain2020synthetic} employed a Generative Adversarial Network (GAN)-based augmentation scheme to synthesize additional images for surface defect detection. \cite{yun2020automated} proposed using the Conditional Variational Auto-Encoder (CVAE) to generate diverse defect images by sampling from the learned latent space. 
\cite{tsai2021auto} proposed a two-stage deep learning scheme for defect detection with only few true defect samples by using CycleGAN. \cite{zhu2017unpaired} to synthesize and annotate defect pixels in an image automatically. \cite{jiang2021casting} first learned to extract the attention maps for each image, based on which the dataset is augmented in a weakly supervised manner. In general, these methods only focus on addressing data augmentation for 2D images. 

Comparing with these potential solutions, our model benefits from the use of a teacher-student framework and multi-scale feature extraction in the 3D point cloud scenario. The reasons lie in that: (i) it allows the student to learn how to specifically extract necessary features from a powerful pretrained teacher network according to the concept of knowledge distillation, which can combat the tendency of the precision loss issue induced by over-valuing generalization performance in few-shot-based models; (ii) by incorporating multi-scale modules into our framework, the student network can learn both coarse and fine features from few training samples, which is more straightforward and unified since it frees synthetically generating scaled or transformed data for training.

\section{Method}
\label{Method}

Fig. \ref{fig:2} illustrates the overview of our method for detecting anomaly with respect to point cloud. Our network is constituted of a teacher-student network pair. The teacher network is first pretrained to characterize the deep features of point cloud for all categories using the feature extraction block adopted from PointNet \cite{qi2017pointnet}. The student network, which is  structured identically to the teacher network, then learns to effectively match the features for a required category of point cloud with the teacher counterparts. Furthermore, we introduce several pooling modules at different scales between the teacher and the student networks, and design a multi-scale loss calculated with all the pooling outputs to fully utilize the features to guide the student network. During inference, we compute the cosine similarity-based metric between the outputs of the teacher and student networks as the anomaly score to evaluate the degree of anomaly. Specifically, given the assumed normal point cloud samples of a certain category, the student network learns to predict the corresponding output feature embedded by the pretrained teacher network during training process (blue arrows in Fig. \ref{fig:2}). During the test process, the anomaly score is calculated based on the similarity between the feature vectors outputted by student and teacher (orange arrows in Fig. \ref{fig:2}). Lower similarity indicates a higher degree of anomaly. \par

The rest of this section is organized as follows. First, we introduce our in detail PointNet-structured teacher-student framework. We then specify our multi-scale optimization strategy. Finally, we present the inference procedure for detecting the test samples.

\subsection{Teacher-Student Structured PointNet for 3D Anomaly Detection}
\noindent	\textbf{Teacher network.} We first detail the training policy for the teacher network.
Formally, we define a set of point cloud data as $\mathcal{M}^{c} = \{ \mathbf{M}^{c}_1, \cdots, \mathbf{M}^{c}_m, \cdots, \mathbf{M}^{c}_{N_c}\}$ for the $c$-th object category with $N_c$ point clouds, where each point cloud $\mathbf{M}^{c}_m = \{\mathbf{p}_1, \cdots, \mathbf{p}_w, \cdots, \mathbf{p}_W \}$ contains $W$ 3D points represented by their coordinates $(x,y,z)$ (i.e., $\mathbf{p}_w \in \mathbb{R}^{3}$). The teacher network is leveraged as a powerful feature extractor to obtain quality deep features for point cloud to facilitate distillation.\par
\noindent \textbf{PointNet.} To achieve this, our teacher network follows the PointNet \cite{qi2017pointnet} to extract features from point cloud. More precisely, we design our teacher network based on the architecture of the PointNet segmentation network \cite{qi2017pointnet}, which is an extension of the basic PointNet with increased model depth and dimension. The general idea of PointNet is to apply a symmetric function on transformed point sets to effectively characterize the property of permutation invariance for the point cloud. Specifically, for the pretraining of the teacher network, an input raw point cloud $\mathbf{M}^{c}_m$ fed to the teacher network is first transformed by an alignment module $A_1$ and then passes through a multilayer perceptron (MLP) encoding module $G_1$ to be encoded as a feature representation $f_1$, as follows:
\begin{equation}
\centering
\label{eq:eq1}
f_1 = G_1(A_1(\mathbf{M}^{c}_m)).
\end{equation}
The features are again aligned and then lifted to higher dimensions with modules $A_2$ and $G_2$ respectively, which is given by
\begin{equation}
\centering
\label{eq:eq2}
f_2 = G_2(A_2(f_1)).
\end{equation}
$f_2$ is the resulting high-dimensional feature embedding. Eventually, the features at different layers are concatenated as a mixture into the final segmentation module. As such, by optimizing the segmentation loss (i.e., cross-entropy), the features in each layer can be gradually refined to capture meaningful features with different scales. 

We pretrain the teacher network on the whole point cloud dataset $\mathcal{M} = \{ \mathcal{M}^{c}|c=1,\cdots,C \}$ to guarantee a complete learning of features for all types of objects. Since the eventual aim of the teacher lies solely in feature extraction, during the training of the teacher-student network, we remove the segmentation module to form the final teacher network. After acquiring a teacher network pretrained on the segmentation task, we next need to determine the corresponding student network to fulfill the task of AD for point cloud.

\noindent	\textbf{Student network.} The target of the student network is to learn to model a specific category (i.e., selected normal category) of point cloud under the guidance of the pretrained teacher network. As suggested in \cite{wang2021student_teacher}, we design a student network structured identically to the teacher network. Such an architecture manifests a two-fold essentiality in terms of: (1) the information loss in distillation (e.g., feature compression/transformation) can be mitigated; (2) the intermediate features can be effectively utilized. We next proceed to explain how to fully exploit the rich features to empower the student learning.

\subsection{Optimization}
\label{subseciton:Optimization}
The optimization of the student network is realized based on the trained teacher network via knowledge distillation. In particular, a core issue for an ideal distillation strategy is the selection of layers to distill knowledge. To completely exploit the advantage of the teacher, we draw inspiration from the previous feature-based distillation schemes \cite{wang2021student_teacher, salehi2021multiresolution} by accumulating the features with different scales to enrich the representation. Given a point cloud sample $\mathbf{M}^{c}_m$ with the label of normal category, we define the output of the teacher network at the position $i$ as ${\rm F}^{\rm T}_{i}(\mathbf{M}^{c}_m)$. Analogous to that of the teacher, the feature regressed by the student network is represented as ${\rm F}^{\rm S}_{i}(\mathbf{M}^{c}_m)$. Both ${\rm F}^{\rm T}_{i}(\mathbf{M}^{c}_m)$ and ${\rm F}^{\rm S}_{i}(\mathbf{M}^{c}_m)$ are given by vectors with the dimensions of $h_i$. In our implementation, we 
employ a triplet of distillation positions (i.e., $i \in \{1,2,3\}$) by setting $(h_1, h_2, h_3) = (128,512,2048)$. Note that the features given by both the teacher and student are processed with the max-pooling operation right before the final output. Hence, each output vector can represent more global features with the corresponding scale to better fulfill AD for objects' structure. 

Previous works \cite{bergmann2020uninformed,du2022unknown} generally adopt $L2$ (or $L1$) loss between the student and teacher output for student training. In the case of our framework, however, we have noticed that such an optimization strategy is prone to poor performance. The reason is that, different from images that are naturally ordered, there is no innate ordering between 3D points, and adopting the $L2$ distance will break the required permutation invariance and manually impose dependencies in the optimization.

To achieve a scale-invariant and order-invariant training strategy, we adopt the cosine-similarity metric between each extracted teacher-student feature pair. For the $i$-th pair, we encourage the angle of the student's output to approximate the corresponding teacher's output by \textit{minimizing} the following objective:
\begin{equation}
\label{eq:eq3}
\mathcal{L}_i(\mathbf{M}^{c}_m)= 1-\frac{{\rm F}^{\rm S}_{i}(\mathbf{M}^{c}_m) \cdot {\rm F}^{\rm T}_{i}(\mathbf{M}^{c}_m)}{{\max}( \Vert {\rm F}^{\rm S}_{i}(\mathbf{M}^{c}_m) \Vert_2 \times \Vert {\rm F}^{\rm T}_{i}(\mathbf{M}^{c}_m) \Vert_2, \epsilon) },
\end{equation}
where $\epsilon$ is a constant introduced to ensure numerical stability. The other constant $1$ is added to fit the objective value into $[0, 2]$ to ease subsequent evaluation without hindering the network optimization. We then average Eq. \ref{eq:eq3} at all the positions with respect to the whole training set $\mathcal{M}^c$ to form the final training loss for our student network:
\begin{equation}
\label{eq:eq4}
\mathcal{L}=\sum_{m=1}^{N_c}\frac{1}{3}\sum_{i=1}^{3}\mathcal{L}_i(\mathbf{M}^{c}_m),i \in \{1, 2, 3\}.
\end {equation}
The pretrained teacher network is kept frozen while optimizing the student network to circumvent the issue of trivial parameterization \cite{deng2022anomaly}. Eventually, \textit{minimizing} Eq. \ref{eq:eq4} motivates
the student to mimic the teacher's behavior in terms of a coarse to fine level of learned representations. Our method requires a category-specific student retraining to allow for detecting anomalous point cloud when the definition of the normal label changes. 

Alg. \ref{alg:alg1} shows the training algorithm for our student network. Importantly, our framework enjoys an essential advantage that the knowledge transferred by the teacher greatly alleviates the sample reliance in training the student network. More precisely, we only use one to five normal samples for the student training to achieve an improved detection performance compared with prior work \cite{masuda2021toward} that involves a large number of training samples to learn features for reconstruction. Such a property also yields a significantly low computational complexity for the student retraining compared with \cite{masuda2021toward}. 

\subsection{Inference}
The trained teacher-student network pair can be jointly leveraged to detect anomalous point cloud. For a test sample $\mathbf{T}^{c_u} \in \mathbb{R}^{w \times 3}$ including $w$ points, our target is to determine whether the unknown label $c_u$ belongs to the normal category $c_n$. Since the student network is trained only to represent the point cloud labeled by $c_n$, the anomalous sample should incur a significantly differed teacher-student feature pair. We thus directly forward $\mathbf{T}^{c_u}$ to the teacher and student paths to respectively calculate the feature vectors. Based on the features, we calculate the anomaly score via Eq. \ref{eq:eq3}. We can then evaluate the degree of anomaly by comparing the computed anomaly score with a predefined threshold. Note that we only use the anomaly score at the final position (i.e., $i = 3$) during the test stage. Please refer to Sec. \ref{sec:ablation} for a detailed ablation study. The inference algorithm is summarized in Alg. \ref{alg:alg2}.

\begin{algorithm}[t]
    \begin{algorithmic}[1]
        \Require{\qin{Training point cloud set $\mathcal{M}^{c} = \{ \mathbf{M}^{c}_1, \cdots, \mathbf{M}^{c}_m, \cdots, \mathbf{M}^{c}_{N_c}\}$, pre-trained teacher network}}
        \Ensure{\qin{Optimized student network}}

                \For{\qin{$\mathbf{M}^{c}_m$ in $\mathcal{M}^{c}$}}
                   \State \qin{Obtain ${\rm F}^{\rm T}_{i}$, ${\rm F}^{\rm S}_{i}$ for $i=1,2,3$}

                   \State \qin{Update student network by \textit{minimizing} Eq. \ref{eq:eq4}}
            
                \EndFor

                \State \qin{\textbf{return} Optimized student network}

    \end{algorithmic}
    \caption{\qin{Training for the student network}}
    \label{alg:alg1}
\end{algorithm}

\begin{algorithm}[t]
    \begin{algorithmic}[1]
        \Require{ \qin{A test point cloud sample, threshold $\mathcal{\tau}$}}
        \Ensure{\qin{Predicted label}}

            \State \qin{Obtain anomaly score  $s$ via Eq. \ref{eq:eq3} $(i=3)$}

            \If{\qin{$s$ \textless $\mathcal{\tau}$}}
                \State \qin{\textbf{return} $0$ (for $normal$)}
            \Else
                \State \qin{\textbf{return} $1$ (for $anomalous$)}
            \EndIf

    \end{algorithmic}
    \caption{\qin{Inference}}
    \label{alg:alg2}
\end{algorithm}

\section{Experiment}
\label{Experiment}
\qin{In this section, we first present the dataset, implementation details, and evaluation metrics used in our experiment. Then, we quantitatively and qualitatively compare our method against the state-of-the-art models. Finally, we perform ablation studies from multiple aspects to provide a deep analysis of our method.}\par
\subsection{Dataset and Implementation Details}
Following the previous point cloud-related literature \cite{klokov2017escape, wang2019dynamic}, we evaluate our method on the large-scale 3D point cloud dataset: ShapeNet-Part dataset \cite{yi2016scalable}. It provides per-point annotation for 16 separate shape categories of objects (airplane, bag, cap, etc). We use the official train/test split with 12,137 samples for training and 2,874 samples for test. Each training and test sample consists of 2,048 randomly sampled  points (i.e., $W=2048$). For the teacher network, we pretrain it for 251 epochs on all the 16 categories in the ShapeNet-Part dataset, following \cite{Pytorch_Pointnet_Pointnet2}. For the student network, we train it only on a selected normal category with very few samples (1 to 5) for 20 epochs. Both the teacher and  student networks are optimized with the ADAM optimizer based on an exponentially decaying learning with the initial value of 1e-3.

\subsection{Evaluation Metric}
We use the area under the Receiver Operating Characteristic (ROC) curve to quantitatively assess our method. The ROC curve reflects the relationship between the True Positive Rate (TPR,
$\overline{TP}/ (\overline{TP}+\overline{FN})$) and False Positive Rate (FPR, $\overline{FP}/ (\overline{FP}+\overline{TN})$) under varying thresholds, where $\overline{TP}, \overline{FN}, \overline{FP}, \overline{TN}$ denote the number of true positive, false negative, false positive, and true negative samples, respectively. The Area Under the Curve (AUC) metric then computes the whole area under such an ROC curve, 
and a larger AUC indicates a higher detection capability.
\begin{table}[]
\caption{\qin{Average AUC over all categories under each method. All the results per category are obtained with the corresponding method trained with five samples.}}
\label{tab:tab1}
\centering
\scalebox{1.2}{
\begin{tabular}{lccc}
\toprule
\qin{\textbf{Category}}             & \qin{{\textbf{\cite{masuda2021toward}}}}  & \qin{\textbf{\cite{floris2022composite}}}  & \qin{\textbf{Ours}} \\ \hline
\qin{Airplane}   & \qin{97.37}                       & \qin{62.26}                  & \qin{\textbf{98.36}} \\
\qin{Bag}        & \qin{61.40}                       & \qin{71.14}                  & \qin{\textbf{99.73}} \\
\qin{Cap}        & \qin{50.64}                       & \qin{51.01}                 & \qin{\textbf{96.61}} \\
\qin{Car}        & \qin{65.53}                       & \qin{62.15}                  & \qin{\textbf{99.08}} \\
\qin{Chair}      & \qin{54.80}                       & \qin{58.59}                  & \qin{\textbf{98.82}} \\
\qin{Earphone}   & \qin{44.66}                       & \qin{68.08}                  & \qin{\textbf{90.86}} \\
\qin{Guitar}     & \qin{78.50}                       & \qin{64.10}                  & \qin{\textbf{97.96}} \\
\qin{Knife}      & \qin{72.40}                      & \qin{83.57}                  & \qin{\textbf{96.16}} \\
\qin{Lamp}       & \qin{56.69}                       & \qin{\textbf{69.69}}                  & \qin{58.36} \\
\qin{Laptop}     & \qin{70.02}                       & \qin{60.01}                  & \qin{\textbf{99.00}} \\
\qin{Motorbike}  & \qin{87.61}                       & \qin{87.39}                  & \qin{\textbf{97.92}} \\
\qin{Mug}        & \qin{41.94}                       & \qin{60.22}                  & \qin{\textbf{98.25}} \\
\qin{Pistol}     & \qin{80.67}                       & \qin{89.12}                  & \qin{\textbf{97.83}} \\
\qin{Rocket}     & \qin{60.32}                       & \qin{82.23}                  & \qin{\textbf{97.80}} \\
\qin{Skateboard} & \qin{53.31}                       & \qin{62.94}                  & \qin{\textbf{97.95}} \\
\qin{Table}      & \qin{79.56}                       & \qin{69.45}                  & \qin{\textbf{87.88}} \\
\qin{Avg. AUC}   & \qin{65.96}                       & \qin{68.87}                  & \qin{\textbf{94.54}} \\
\bottomrule
\end{tabular}}
\end{table}

\begin{table*}
\centering
\caption{The first column depicts the target category for training (i.e., the normal category). When a target category is determined, the other 15 categories are treated as anomalous categories. Each row denotes the quantitative comparison with respect to category-wise AUC ($\%$) metric against \cite{masuda2021toward} on each category with a different number of training samples (i.e., 1, 3, 5). For example, when ``Airplane’’ is the target category, 341 normal airplane tests and another 2,533 anomalous tests are used for testing. The digits to the left and right of ``$\pm$’’ denote the average AUC and the standard derivation for 10 runs with randomly selected training samples, respectively. The best result of each row is in bold.}
\label{tab:tab2}
\scalebox{1.2}{
\begin{tabular}{l|ccc|ccc}
\toprule
\multicolumn{1}{c|}{\multirow{2}{*}{\textbf{Category (\#tests)}}} & \multicolumn{3}{c|}{\begin{tabular}[c]{@{}c@{}} {\textbf{Reconstruction-based method}} \\{\textbf{\cite{masuda2021toward}}}\end{tabular}} & \multicolumn{3}{c}{\begin{tabular}[c]{@{}c@{}} {\textbf{Knowledge-distillation-based method}} \\ {\textbf{(Ours)}} \end{tabular}}                                                 \\
\multicolumn{1}{c|}{}                                   & \textbf{1 sample}   & \textbf{3 samples}  & \textbf{5 samples}         & \textbf{1 sample}         & \textbf{3 samples}         & \textbf{5 samples}        \\ \hline
Airplane (341)                                               & 87.60 $\pm$ 5.99    & 96.07 $\pm$ 1.07    & 97.39 $\pm$ 0.48           & 97.41 $\pm$ 1.01          & \textbf{98.59 $\pm$ 0.18} & 98.29 $\pm$ 0.53          \\
Bag (14)                                                     & 47.09 $\pm$ 8.73    & 52.41 $\pm$ 5.86    & 58.70 $\pm$ 5.79           & 96.88 $\pm$ 3.73          & 98.23 $\pm$ 2.13          & \textbf{99.94 $\pm$ 0.08} \\
Cap (11)                                                     & 38.71 $\pm$ 5.70    & 45.22 $\pm$ 3.82    & 46.31 $\pm$ 6.38           & 90.96 $\pm$ 5.54          & \textbf{94.96 $\pm$ 3.05} & 94.13 $\pm$ 2.94          \\
Car (158)                                                     & 62.28 $\pm$ 3.27    & 64.12 $\pm$ 2.06    & 65.14 $\pm$ 1.81           & 99.33 $\pm$ 0.27          & \textbf{99.36 $\pm$ 0.26} & 99.31 $\pm$ 0.29          \\
Chair (704)                                                  & 49.20 $\pm$ 4.27    & 53.38 $\pm$ 2.96    & 55.38 $\pm$ 1.29           & 95.16 $\pm$ 2.20          & 98.54 $\pm$ 0.64          & \textbf{98.72 $\pm$ 0.18} \\
Earphone (14)                                               & 38.78 $\pm$ 7.61    & 45.36 $\pm$ 7.26    & 43.64 $\pm$ 3.21           & 81.97 $\pm$ 23.45         & \textbf{91.31 $\pm$ 3.07} & 90.19 $\pm$ 2.22          \\
Guitar (159)                                                  & 71.75 $\pm$ 3.45    & 76.13 $\pm$ 3.74    & 77.59 $\pm$ 2.57           & \textbf{98.65 $\pm$ 0.54} & 97.66 $\pm$ 1.13          & 98.39 $\pm$ 0.69          \\
Knife (80)                                                   & 66.46 $\pm$ 4.20    & 70.49 $\pm$ 2.31    & 71.79 $\pm$ 0.91           & 95.18 $\pm$ 1.75          & 95.33 $\pm$ 2.22          & \textbf{96.72 $\pm$ 1.08} \\
Lamp (286)                                                    & 53.09 $\pm$ 5.02    & 58.68 $\pm$ 2.55    & \textbf{62.20 $\pm$ 3.68}  & 56.13 $\pm$ 6.16          & 60.76 $\pm$ 8.34          & 61.22 $\pm$ 5.84          \\
Laptop (83)                                                 & 67.36 $\pm$ 4.52    & 69.08 $\pm$ 2.24    & 70.24 $\pm$ 2.58           & \textbf{98.89 $\pm$ 0.16} & 98.78 $\pm$ 0.32          & 98.69 $\pm$ 0.29          \\
Motorbike (51)                                               & 82.55 $\pm$ 2.75     & 87.66 $\pm$ 1.28    & 88.09 $\pm$ 2.56           & 90.63 $\pm$ 26.57         & 99.27 $\pm$ 0.72          & \textbf{99.35 $\pm$ 0.60} \\
Mug (38)                                                     & 44.04 $\pm$ 5.67    & 46.94 $\pm$ 2.44    & 48.55 $\pm$ 3.82           & 99.60 $\pm$ 0.49          & \textbf{99.73 $\pm$ 0.37} & 99.35 $\pm$ 0.57          \\
Pistol (44)                                                 & 64.85 $\pm$ 4.70    & 71.77 $\pm$ 4.41    & 78.57 $\pm$ 4.44           & \textbf{99.08 $\pm$ 0.66} & 98.86 $\pm$ 0.51          & 98.80  $\pm$ 0.48         \\
Rocket (12)                                                  & 55.25 $\pm$ 2.97    & 55.95 $\pm$ 5.69    & 59.67 $\pm$ 5.05           & \textbf{96.63 $\pm$ 1.44} & 95.73 $\pm$ 2.01          & 96.23 $\pm$ 2.93          \\
Skateboard (31)                                             & 47.84 $\pm$ 5.19    & 57.56 $\pm$ 5.08    & 57.29 $\pm$6.62            & \textbf{96.16 $\pm$ 1.20} & 95.64 $\pm$ 1.06          & 96.07 $\pm$ 1.11          \\
Table (848)                                                   & 47.39 $\pm$ 14.48   & 61.97 $\pm$ 14.66   & 78.78 $\pm$ 4.56           & 83.46 $\pm$ 8.17          & 89.77 $\pm$ 2.34          & \textbf{90.40 $\pm$ 1.05} \\
Avg. AUC                                                 & 57.77               & 63.30               & 66.21                      & 92.26                     & 94.53                     & \textbf{94.74}            \\
Avg. of Std. Dev.                                      & 5.53                & 4.21                & 3.48                       & 5.21                      & 1.77                      & \textbf{1.30}             \\ \bottomrule
\end{tabular}}
\end{table*}

\begin{figure*}[tb]
     \centering
     \includegraphics[width=1.0\linewidth]{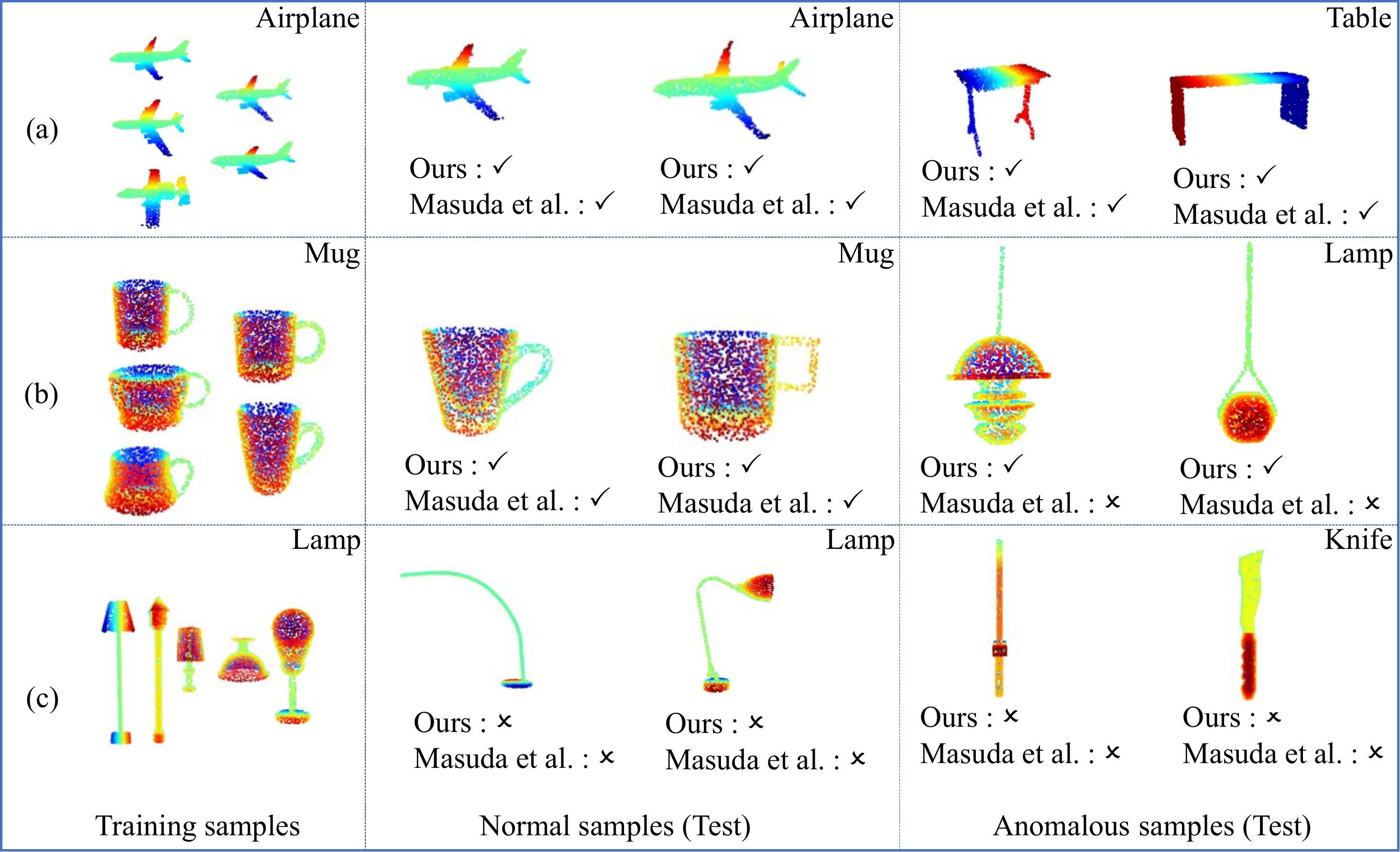}
     \caption{Qualitative comparisons of three selected categories for normal and anomalous samples. The columns from left to right denote training samples, normal test samples, and anomalous test samples, respectively. The training/test category for the cell is explained in the upper-right corner. (a) $\sim$ (c) show three detection results for each assumed normal category in the left-most cell. We use ``\checkmark’’ and ``$\times$’’ to respectively indicate the detection success or failure.}
     \label{fig:3}
\end{figure*}

\subsection{Comparative Results}
\noindent	\textbf{Quantitative Results.} We first report the quantitative results in terms of detection accuracy. We compare our method against the closest state of the arts: \cite{masuda2021toward,floris2022composite} that handle the same problem as ours. We give a comprehensive evaluation of detection performance, in which the student network is retrained on each category for different normal category assumptions (i.e., when a certain category is treated as normal, the remaining 15 categories are treated as anomalous). To manifest the ability of our method in handing few samples, during the training for each category selected as normal, we randomly sample \qin{5} point clouds from the corresponding category, and then train \cite{masuda2021toward,floris2022composite} on the same sampled data as ours to make the training protocol for both methods consistent. \textit{Note that the comparison is, however, not fair because (i) the \cite{masuda2021toward,floris2022composite} \qin{are} not originally designed for dealing with few samples as ours; (ii) our framework involves a teacher pretraining stage, whereas \cite{masuda2021toward,floris2022composite} \qin{do} not.
Therefore, the comparative results are only focused on presenting the efficiency of our teacher-student framework in modeling the required normal categories with few samples.}

The results are summarized in Tab. \ref{tab:tab1}.
It can be clearly observed that the proposed method outperforms \cite{masuda2021toward,floris2022composite} in detection accuracy by a large margin for almost all categories and sample number settings. Particularly, for \cite{masuda2021toward}, this is because the limited training samples considerably degenerate the reconstruction strength of the VAE model in \cite{masuda2021toward},
causing the chamfer distance between normal and anomalous categories to be less sensitive to fulfill the detection.
\qin{As for \cite{floris2022composite}, although it is designed to learn  local spatial relations for points, the domain-dependent geometric transformation it utilizes to realize self-supervision is originally devised for images, which may not suit well when dealing with 3D point clouds. In addition, the few sample training further imposes challenges on detection.}

In general, unlike \cite{masuda2021toward,floris2022composite}, our method does not show noticeable reliance on the training sample number. We expect this to be attributed to the fact that the pretrained teacher network well compensates for the lack of samples for student training, which also justifies the large AUC gap between ours and \cite{masuda2021toward,floris2022composite}.

\noindent	\textbf{Qualitative Results.} We next qualitatively investigate our method to provide more insights. Fig. \ref{fig:3} visualizes several detection results for some categories on both normal and anomalous test samples. It can be seen in Fig. \ref{fig:3} that the shape diversity within each category has a huge impact on the detection accuracy. The more diversified the object shapes are, the more challenging the detection will be. For example, both our method and \cite{masuda2021toward} perform well on the category of ``Airplane’’, where almost all the training/test samples are similar (Fig. \ref{fig:3}(a)). However, in the case of ``Lamp’’ (Fig. \ref{fig:3}(c)), because the training and test samples have remarkably separate shapes (e.g., width or straightness of the lamp body), both methods tend to fail, which is also reflected in the low AUC in Tab. \ref{tab:tab1} (10th row). 

Furthermore, reconstruction-based method can neglect some fine object features and only concentrates on the global shape. For example, when dealing with the category ``Mug’’, the method by \cite{masuda2021toward} cannot well capture the thin line above the lamp (last column in Fig. \ref{fig:3}(b)), wrongly classifying the lamp sample as normal. Contrarily, our method handles such a case successfully because we employ the multi-scale training loss to extract features at different levels.

\begin{figure}[tb]
     \centering
     \includegraphics[width=1\linewidth]{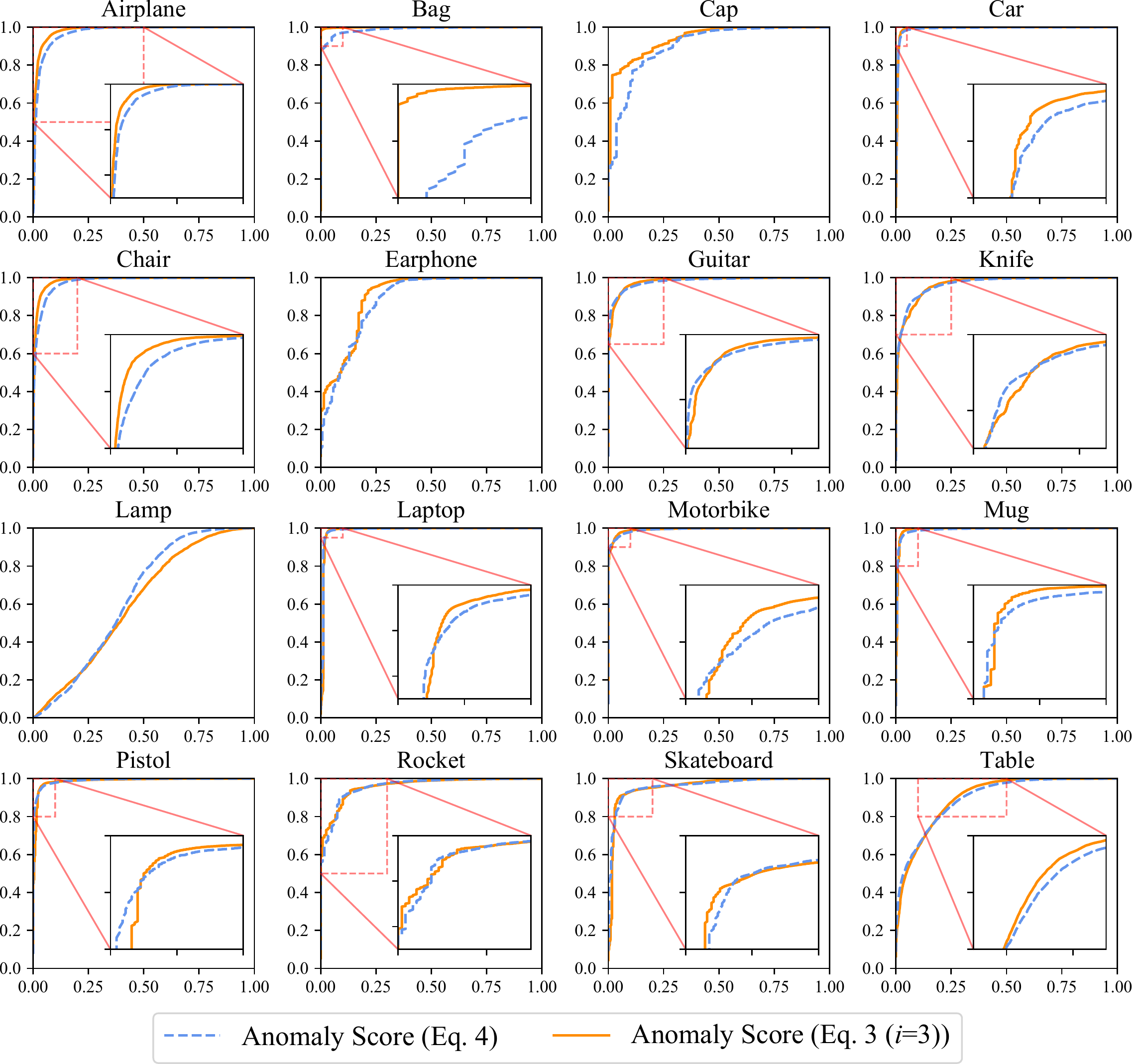}
     \caption{ROC curve of Anomaly Score at different designed scales. Test performance in each category was evaluated by the final scale and multi-scale anomaly score separately. }
     \label{fig:4}
     \end{figure}

\begin{figure}[tb]
     \centering
     \includegraphics[width=1\linewidth]{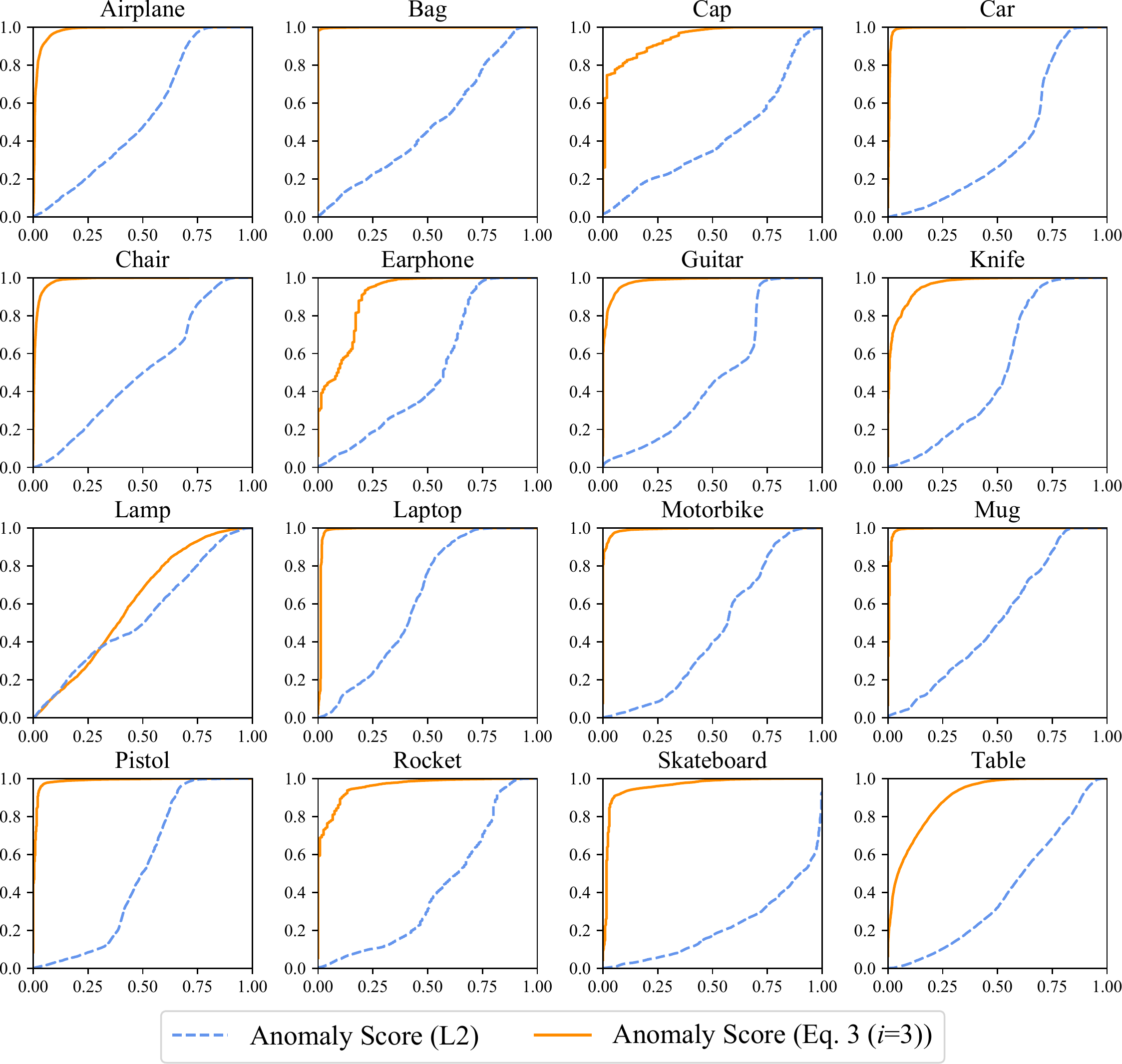}
     \caption{ROC curve of Anomaly Score at different designed losses. Test performance in each category was evaluated by using the $L2$ and cosine similarity anomaly score separately.}
     \label{fig:5}
     \end{figure}

\begin{figure}[tb]
     \centering
     \includegraphics[width=1\linewidth]{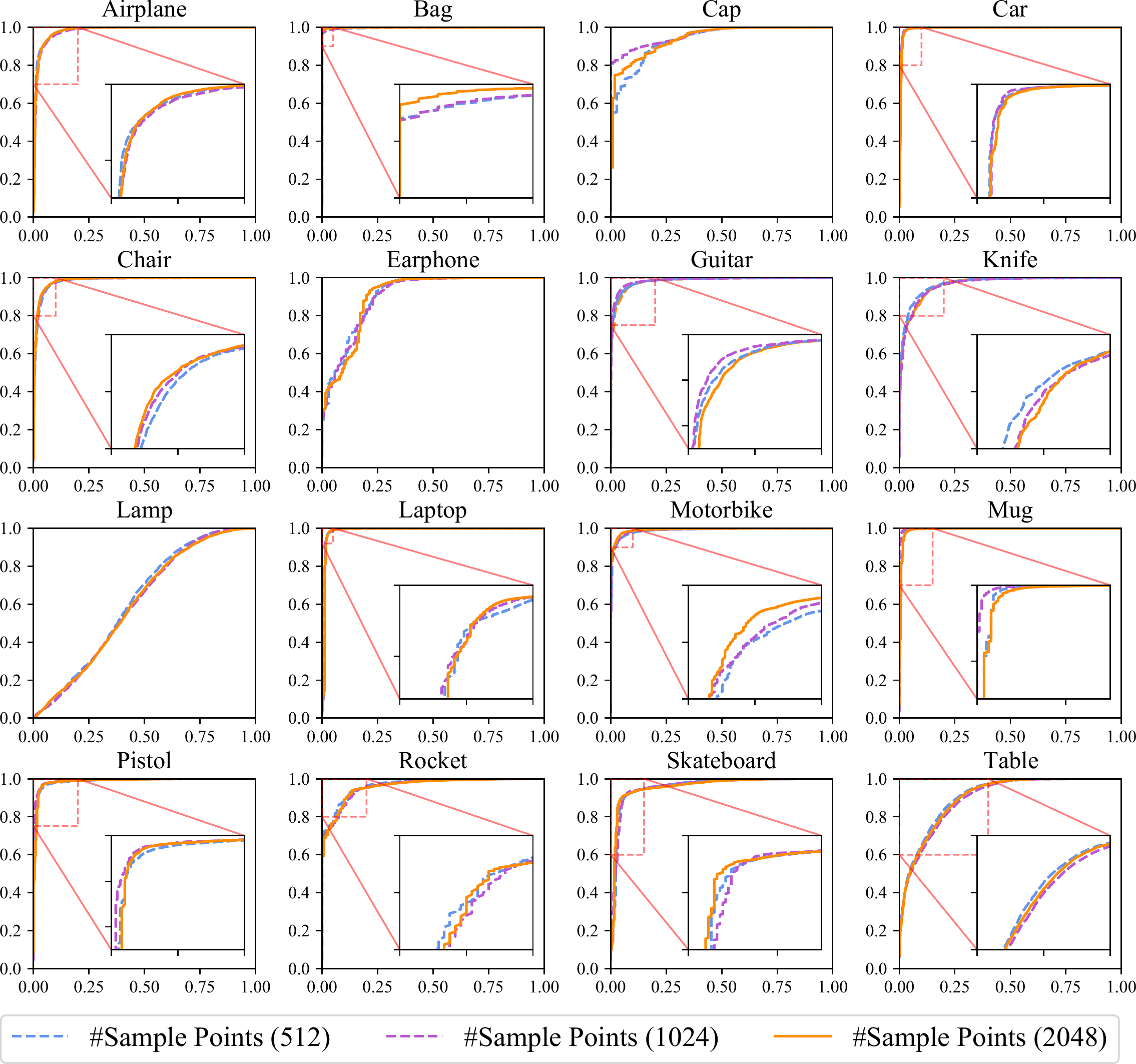}
     \caption{Comparison of results at different sample points. Test performance in each category was evaluated by using 512, 1,024, and 2,048 sample points separately.}
     \label{fig:6}
     \end{figure}

\begin{table*}[]
\caption{\qin{Average AUC over all categories under different student structures when training with five samples. The first row shows the dimensions of the student network at $h_1, h_2, h_3$ from left to right, except in the second column where $h_2$ is removed. The teacher network follows 128-512-2048 at $h_1, h_2, h_3$.}}
\label{tab:tab3}
\centering
\scalebox{1}{
\begin{tabular}{cccccc}
\toprule
\qin{\textbf{Student structure}}   & \qin{\textbf{128-2048}} & \qin{\textbf{128-32-2048}} & \qin{\textbf{128-64-2048}} & \qin{\textbf{128-256-2048}} & \qin{\textbf{128-512-2048}} \\ 
\qin{\textbf{(Num. of distillation positions) }}& \qin{\textbf{(2)}}    & \qin{\textbf{(2)}}       & \qin{\textbf{(2)}}       & \qin{\textbf{(2)}}       & \qin{\textbf{(3)}}       \\\hline
\qin{Avg. AUC }& \qin{93.93}    & \qin{94.03}       & \qin{94.09}       & \qin{94.24 }       & \qin{\textbf{94.54}}       \\ \bottomrule
\end{tabular}}
\end{table*}

\subsection{Ablation Studies}
\label{sec:ablation}

To provide a deeper understanding of our model, we here
analyze the effects of our model regarding the following aspects:

\noindent	\textbf{Data Sampling for Training.}
Our proposed model requires only one to five randomly chosen normal samples for training, which can impose the training stochasticity that stems from the quality of the sampled point clouds. To analyze the underlying effect, we conduct 10 runs in Tab. \ref{tab:tab2} by sampling different point clouds.
For fair comparisons, all the experimental settings for training and testing \cite{masuda2021toward} are kept the same. As indicated by the average standard deviation in Tab. \ref{tab:tab2} (last row), our method is less dependent than \cite{masuda2021toward} on sample number (i.e., the value is much smaller). Furthermore, for some categories, even one training sample yields the highest AUCs (5th column in Tab. \ref{tab:tab2}), especially regarding the cases of more than one sample. In particular, using three or five results shows stable performance. We thus set the sample number to five for all the following ablations since it performs the best.

\noindent	\textbf{Scales of Features for Test.} 
Although we employ multi-scale training loss to facilitate the feature matching at different positions, the issue of which feature contributes the most to the test performance remains unexplored. We thus evaluate the detection performance by using multi-scale and single-scale anomaly scores for inference to respectively examine the contribution of different positions. We follow Eq. \ref{eq:eq4} to calculate the multi-scale anomaly score. For the single-scale case, we compute the anomaly score at the final scale ($i=3$) via Eq. \ref{eq:eq3}. We can observe in Fig. \ref{fig:4} that single- and multi-scale cases generally perform comparably, with the single-scale anomaly score performing marginally better. Motivated by this observation, we simply follow the single-scale test policy in all the experiments.


\noindent	\textbf{Different Types of Anomaly Scores for Test.}
To show how the different anomaly scores affect the performance, we compare the designed anomaly score (i.e., cosine similarity) against the $L2$ metric during inference. As shown in Fig. \ref{fig:5}, the cosine similarity-based anomaly score significantly outperforms the $L2$ metric for almost all categories, demonstrating  
its capacity in capturing scale-invariant features for detection.

\noindent	\textbf{Number of Points.}
We train and test our model using different numbers of sample points for the point cloud to
investigate the impact. We can observe in Fig. \ref{fig:6} that more sampled points yield a stronger representation for object shape, thus leading to higher detection performance in many cases. However, the change of detection accuracy does not vary greatly.

\noindent	\textbf{\qin{Student Network Structure.}}
\qin{As suggested in \cite{wang2021student_teacher}, our student network adopts the identical structure as the teacher network to facilitate distillation. To examine the validity of such a design, we first vary the student architecture by reducing the dimensions at each distillation position (i.e., $(h_1, h_2, h_3)$ in Sec. \ref{subseciton:Optimization}), and then retrain the student network for each scenario. Note that the distillation is only enforced if the student and the teacher have the same dimensions. Tab. \ref{tab:tab3} shows the results. We can observe that a simplified student network tends to degrade the performance. This is because fewer distillation positions may weaken the capability of the student network to capture multi-scale features for detection. Motivated by this observation, our student network is designed to follow the same architecture as the teacher network in our implementation.}


\section{Conclusion}
In this paper, we have proposed a novel teacher-student network for 3D point cloud anomaly detection with only few normal samples. We also introduce a multi-scale loss function to exploit the potentially beneficial features during training. Extensive experimental results on a large-scale point cloud dataset demonstrate that our method outperforms the state of the art with respect to detection accuracy and dependence on the number of samples. The ablation study further justifies our method.

Our method involves a major limitation if the selected normal category includes highly diversified object shapes. This is mainly because the randomly sampled few training data generally do not cover the diverse shape modes in each category. In the future, we would like to introduce a sampling strategy to explicitly encourage a rich shape coverage within the few training samples.

\section{Acknowledgement}
This work is supported by JSPS KAKENHI Grant Number JP20K19568. We also acknowledge China Scholarship Council (CSC) for funding the first author under Grant Number 202108330097.

\ifCLASSOPTIONcaptionsoff
  \newpage
\fi



%
\bibliographystyle{abbrv}
\bibliography{reference}

%






\end{document}